%% file: main.tex
\documentclass[11pt,a4paper]{article}
\usepackage[hyperref]{emnlp2020}
\usepackage{times}
\usepackage{latexsym}

\usepackage{graphicx}
\usepackage{tablefootnote}
\usepackage{comment}
\usepackage{xcolor,colortbl}
\definecolor{lightblue}{rgb}{0.68, 0.85, 0.9}
\definecolor{lavender}{rgb}{0.9, 0.9, 0.98}
\definecolor{lightyellow}{rgb}{1.0, 1.0, 0.88}
\definecolor{magicmint}{rgb}{0.67, 0.94, 0.82}
\definecolor{palepink}{rgb}{0.98, 0.85, 0.87}
\definecolor{bubbles}{rgb}{0.91, 1.0, 1.0}

\usepackage{microtype}

\input{math_commands}

\usepackage{xcolor}
\usepackage{eqnarray}
\usepackage{booktabs}
\usepackage{multirow}
\usepackage{fixltx2e}
\usepackage{wrapfig}
\usepackage{subfiles}
\usepackage{url}

\usepackage[nameinlink]{cleveref}
\crefformat{section}{\S#2#1#3} 
\crefformat{subsection}{\S#2#1#3}
\crefformat{subsubsection}{\S#2#1#3}

\aclfinalcopy 
\def\aclpaperid{2733} 

\newcommand*{\affaddr}[1]{#1} 
\newcommand*{\affmark}[1][*]{\textsuperscript{#1}}
\newcommand*{\email}[1]{\text{#1}}

\title{DAGA: Data Augmentation with a Generation Approach for Low-resource Tagging Tasks}

\author{Bosheng Ding\thanks{\;  Equal contribution. Bosheng Ding and Linlin Liu are under the Joint PhD Program between Alibaba and Nanyang Technological University.}\affmark[~~12]~~~ Linlin Liu\footnotemark[1]\affmark[~~12]~~~ Lidong Bing\affmark[2]~~~ Canasai Kruengkrai\thanks{\; Work done while at DAMO Academy, Alibaba Group}\affmark[~~3] \\\textbf{Thien Hai Nguyen\affmark[2]~~~
Shafiq Joty\affmark[1]~~~ Luo Si\affmark[2]~~~ Chunyan Miao\affmark[1]}\\
\affaddr{\affmark[1]Nanyang Technological University, Singapore}\\
\affaddr{\affmark[2]DAMO Academy, Alibaba Group}
\affaddr{~~\affmark[3]National Institute of Informatics, Japan}\\
\email{\small{\tt \{bosheng.ding, linlin.liu, l.bing, thienhai.nguyen, luo.si\}@alibaba-inc.com}}\\
\email{\small{\tt canasai@nii.ac.jp~~~}}
\email{\small{\tt\{srjoty, ascymiao\}@ntu.edu.sg}}}

\begin{document}
\maketitle

\begin{abstract}
Data augmentation techniques have been widely used to improve machine learning performance as they enhance the generalization capability of models. In this work, to generate high quality synthetic data for low-resource tagging tasks, we propose a novel augmentation method with language models trained on the linearized labeled sentences. Our method is applicable to both supervised and semi-supervised settings. For the supervised settings, we conduct extensive experiments on named entity recognition (NER), part of speech (POS) tagging and end-to-end target based sentiment analysis (E2E-TBSA) tasks. For the semi-supervised settings, we evaluate our method on the NER task under the conditions of given unlabeled data only and unlabeled data plus a knowledge base.
The results show that our method can consistently outperform the baselines, particularly when the given gold training data are less.\footnote{Our code is available at \url{https://ntunlpsg.github.io/project/daga/}} 
\end{abstract}

\section{Introduction}
\label{sec:intro}

A large amount of training data is often essential for neural model performance, especially, for large networks. Having more training data can help reduce overfitting and improve model robustness. However, preparing a large amount of annotated data is usually costly, labor intensive and time-consuming. Data augmentation \cite{Simard1998} is a useful technique for synthetic data generation, which is widely used in computer vision \citep{fawzi2016adaptive,d2017bridging,wang2019survey} and speech \citep{schluter2015exploring,ko2017study}. 

However, due to the complexity of language, it is more challenging to apply data augmentation techniques to natural language processing (NLP). Unlike computer vision and speech, where handcrafted rules (such as rotation, cropping, masking, etc.) can be easily applied to transform original data, it is difficult to generalize such rules for languages. Although simple distortion usually does not change the semantics of visual information, deleting or replacing a single word could completely change the meaning of the sentence.

\begin{figure}[t!]
\centering
\includegraphics[scale=0.34]{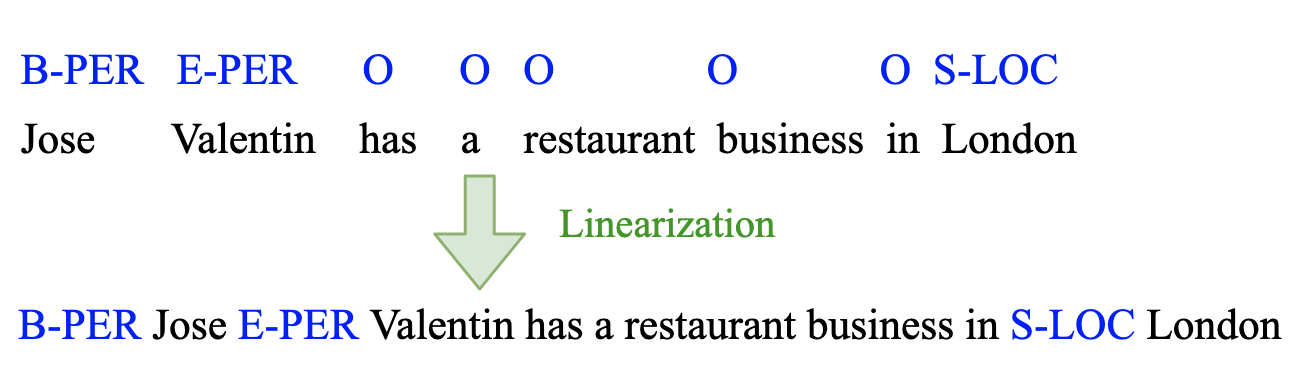}
\caption{An example of labeled sentence linearization. All words and their tags are paired up by inserting tags before (or after) the words (\emph{O} tags removed).}
\label{fig:linearization}
\end{figure}

One successful method for data augmentation in NLP is \emph{back translation} \citep{sennrich-etal-2016-improving,fadaee2017data,dong-etal-2017-learning,wei2018fast}, where a translation model is used to translate monolingual sentences from target language to source language to generate synthetic parallel sentences. Other successful methods include: systematically reordering the dependents of some nodes in gold data to generate synthetic data for dependency parsing \citep{wang2016galactic}, leveraging knowledge base for question generation \citep{serban2016generating} and using simulation-based approach to generate a set of prerequisite toy tasks for QA \citep{weston2015towards}. Besides, synonym replacement, random deletion/swap/insertion, generation with VAE or pre-trained language models are also used in some NLP tasks \citep{kobayashi2018contextual,wei2019eda,anaby2020not,raille2020fast,kumar2020data}, but mainly for translation and classification tasks. 

Compared with the above-mentioned downstream tasks like translation and classification, sequence tagging is more fragile when it is confronted with data augmentation noises due to the finer granularity of the (token-level) task. Annotating unlabeled data with a weak tagger, leveraging aligned bilingual corpora to induce annotation and synonym replacement are three attempted data augmentation methods for sequence tagging \citep{shang2018automated,yarowsky2001inducing,mathew2019biomedical}. Weakly labeled data will inevitably introduce more noise. Note that annotating unlabeled data with a weak tagger requires in-domain data and in-domain knowledge, otherwise it may suffer from domain-shift problem \cite{bari2020multimix}. Leveraging aligned bilingual corpora requires additional resources, which may not be available for low resource languages. Synonym replacement often relies on additional knowledge, e.g., WordNet \citep{miller1995wordnet}, which is a manually designed dictionary that may have low coverage (or not available) for low-resource languages.

In this work, we investigate data augmentation with a generation approach for sequence tagging tasks. We first linearize the labeled sentences as shown with an example in Figure~\ref{fig:linearization}. Then a language model (LM) is trained on the linearized data and used to generate synthetic labeled data. Unlike employing weak taggers to label unseen data, our method unifies the processes of sentence generation and labeling using a LM. 
Concretely, a word and its tag in a pair (e.g., ``B-PER Jose'') are trained to be generated together, in which the tag-word pairs with high probability will be chosen by the LM during generation (\cref{subsec:lm_and_gen}). Our method does not require additional resources like WordNet. Nevertheless, if unlabeled data or knowledge bases are available, our method is also flexible to utilize these resources with a simple but effective conditional generation technique (\cref{subsec:conditional_generation}). 

Although some recent work \citep{anaby2020not,raille2020fast,kumar2020data} also leverages LM for data augmentation, their methods are conditioned on sentence-level tags to generate or modify training data, hence applicable to classification tasks exclusively. To the best of our knowledge, we are the first to utilize generative language models to generate fine-grained synthetic data from scratch for sequence tagging tasks, which introduces a new paradigm for data augmentation in NLP. Furthermore, our method does not rely on large pre-trained models and in fact, it employs a simple one-layer recurrent language model \citep{sundermeyer2012lstm}, which is more convenient to train. Our method demonstrates  encouraging performance when trained on just a few thousand sentences (\cref{sec:exp}).

To verify the effectiveness of our method, we conduct extensive experiments on different sequence tagging tasks, including named entity recognition (NER), part-of-speech (POS) and end-to-end target based sentiment analysis (E2E-TBSA). Our method consistently outperforms the baseline methods in both supervised and semi-supervised settings. Different from the baseline methods, our method generates novel synthetic data from scratch, and thus introduces more diversity to reduce overfitting. For the semi-supervised settings, our method demonstrates strong ability to exploit useful information from unlabeled data and knowledge base.

\section{Background}
\label{sec:background}

\paragraph{Named Entity Recognition (NER)}
Named entities refer to phrases that are names of persons, organizations and locations, etc. in text. For example, ``\emph{[ORG U.N.] official [PER Ekeus] heads for [LOC Baghdad]} ".
Named entity recognition is an important task of information extraction and it aims to locate and classify named entities in text into the predefined types \citep{mikheev1999named,sang2003introduction,li2020survey}. It is a challenging task for two reasons  \citep{lample2016neural}: 1) in most languages and domains, the amount of manually labeled training data for NER is limited; 2) it is difficult to generalize from this small sample of training data due to the constraints on the kinds of words that can be names.

\paragraph{Part-of-Speech (POS) Tagging}
Part-of-speech tagging consists of assigning a tag that represents a grammatical class to each word in a given sentence. It is a critical component of most NLP systems and is fundamental to facilitate downstream tasks such as syntactic parsing \citep{schutze1993part} and opinion analysis \citep{liu-etal-2015-fine}.
The current state-of-the-art POS taggers can achieve over 97.80\% accuracy on PTB-WSJ \citep{akbik2018contextual,bohnet2018morphosyntactic} and yield over 96.50\% average test accuracy across 21 high-resource languages in UD 1.2 \citep{heinzerling2019sequence}. However, one of the problems with current POS taggers is that their accuracy can decrease significantly on low-resource languages and rare words \citep{plank-etal-2016-multilingual,yasunaga2018robust}.

\paragraph{Target Based Sentiment Analysis}
The target based sentiment analysis is a fundamental task of sentiment analysis and it aims to detect the opinion targets in sentences and predict the sentiment polarities over the targets \citep{liu-etal-2015-fine,chen-EtAl:2017:EMNLP20171,li2018aspect,li2019unified}.  For example, \emph{``\textbf{USB3 Peripherals} are noticeably less expensive than the \textbf{ThunderBolt ones}"}. In this sentence, two opinion targets were mentioned, namely ``\emph{\textbf{USB3 Peripherals}}" and ``\emph{\textbf{ThunderBolt ones}}" and the user expresses a positive sentiment over the first, and a negative sentiment over the second. \citet{li2019unified,li2019bert-e2e-absa} propose an end-to-end solution (E2E-TBSA) of TBSA, which converts TBSA to a tagging task, and aims to solve the two subtasks (i.e. target detection and sentiment classification) in a unified manning by predicting unified tags. For example, the tag ``B-POS'' indicates the beginning of a target with positive sentiment. So after annotation, the above example becomes \emph{``[B-POS USB3] [E-POS Peripherals] are noticeably less expensive than the [B-NEG ThunderBolt] [E-NEG ones]"}.

\section{Proposed Method}
\label{sec:method}

We propose a novel data augmentation method for sequence tagging tasks. We first linearize labeled sentences, and then train a language model to learn the distribution of words and tags from the linearized sequences for generating synthetic training data. A conditional generation technique is also proposed to exploit unlabeled data and knowledge bases when they are available.

\subsection{Labeled Sentence Linearization}
\label{ssec:sentence_linearization}
We first perform sentence linearization to convert labeled sentences into linear sequences, so that language models can be used to learn the distribution of words and tags in gold data. As shown in Figure~\ref{fig:linearization}, tags are inserted before the corresponding words during linearization and thus treated as modifiers of these words. For the tasks with frequent \emph{O} tags,  e.g., NER and E2E-TBSA \citep{li2019unified}, we remove such tags from the linearized sequences. Similarly, we can also insert tags after the corresponding words.

After sentence linearization, we add special tokens \emph{[BOS]} and \emph{[EOS]} to the beginning and the end of each sentence,  respectively. 
These special tokens are used to facilitate model training and data generation by marking the sentence boundaries.

\subsection{Language Modeling and Data Generation}
\label{subsec:lm_and_gen}
\begin{figure}[t!]
\centering
\includegraphics[scale=0.38]{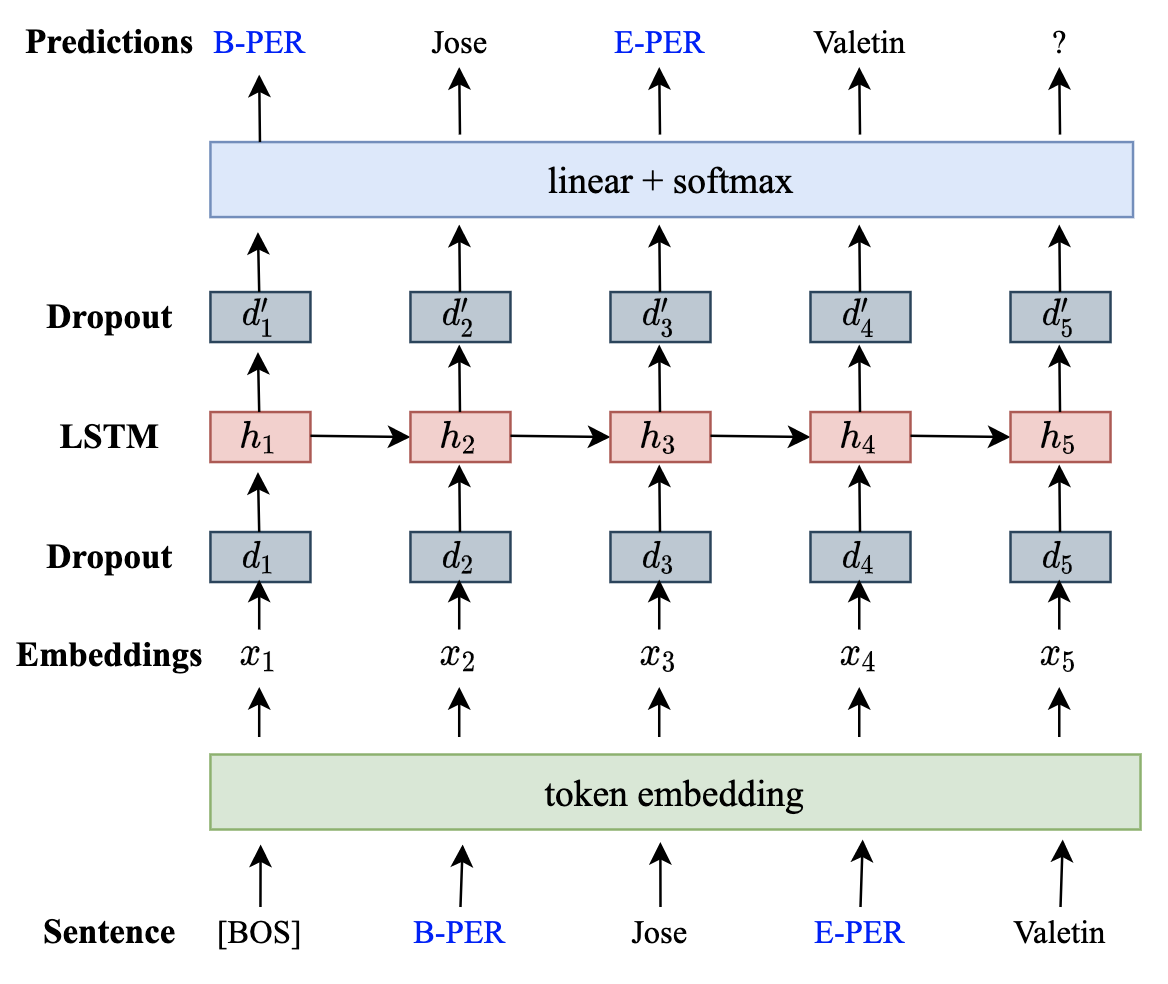}
\caption{Language model architecture with LSTM.}
\label{fig:rnnlm}
\end{figure}

After linearizing labeled sentences, language models can be used to learn the distribution of words and tags.  More specifically, we use a one-layer LSTM recurrent neural network language model (RNNLM) in our method, which is similar to the model proposed by \citet{sundermeyer2012lstm}. The architecture of our RNNLM is shown in Figure~\ref{fig:rnnlm}.

\paragraph{Language Modeling}

We train RNNLM by maximizing the probability for next token prediction. Given a sentence, we first feed the sequence of tokens $(w_1, w_2, \dots, w_N)$ into the embedding layer to lookup the corresponding embeddings $(\vx_1, \vx_2, \dots, \vx_N)$, where $N$ is the sequence length. A dropout layer is applied to each token embedding $\vx_t$ to generate $\vd_t = \dropout(\vx_t)$. Then we feed $(\vd_1, \vd_2, \dots, \vd_N)$ into LSTM to produce hidden state $\vh_t = \LSTM(\vd_t,\vh_{t-1})$ at each position $t$. Another dropout layer is applied to hidden states to compute $\vd^\prime_t = \dropout(\vh_t)$.

\begin{figure*}[t!]
\centering
\includegraphics[scale=0.45]{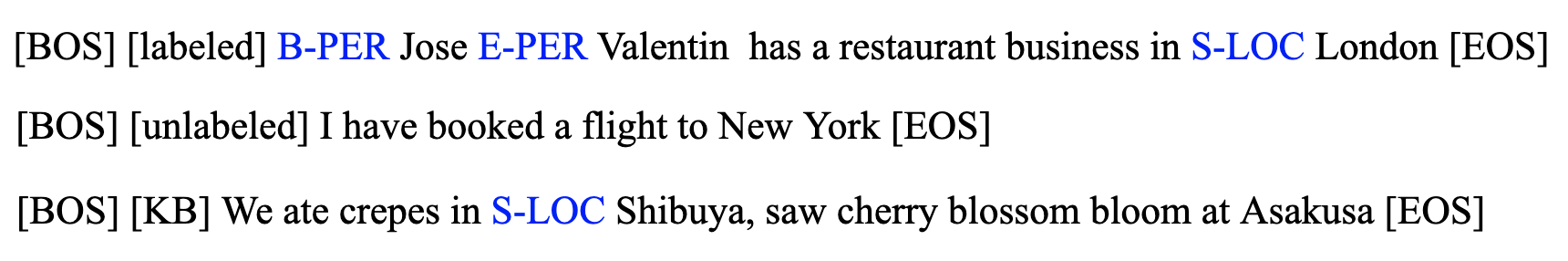}
\caption{An example of conditional generation. The first sequence is from gold NER data. The second is from unlabeled data, so no labels. The third is labeled by knowledge base matching, where \emph{Asakusa} cannot be labeled due to incomplete knowledge coverage.}
\label{fig:conditional_generation}
\end{figure*}

Finally, a linear+softmax layer is used to predict the next token in the sequence. Assuming the index of token $w_t$ in the vocabulary is $i^*$, we have the training objective in Eq.~\ref{eq:train_objective}:
\begin{eqnarray}
    \vs_{t-1} = \mM^\mathsf{T}\vd^\prime_{t-1} \\
    p_{\theta}(w_t|w_{<t}) = \frac{\exp(\vs_{t-1,i^*})}{\sum_{i=1}^V \exp(\vs_{t-1,i})}
    \label{eq:word_prediction} \\
    p(w_1, w_2, \dots, w_N) = \prod_{t=1}^N p_{\theta}(w_t|w_{<t})
    \label{eq:train_objective}
\end{eqnarray}
where $V$ is the size of vocabulary, $\mM \in \mathbb{R}^{r \times V}$ is a learnable weight matrix with $r$ being the dimension of LSTM hidden states, and $\vs_{t-1,i}$ is the $i$-th element of $\vs_{t-1}$.

\paragraph{Generation}
After training the RNNLM, we can use it to generate synthetic training data for tagging tasks. During generation, only the \emph{[BOS]} token is fed into RNNLM, and the following tokens are sampled based on the probabilities computed by Eq.~\ref{eq:word_prediction}. Given \emph{[BOS]}, the sentence is generated autoregressively one at a time, where the token generated in the previous step is taken as input to generate the next one.

As shown in Eq.~\ref{eq:word_prediction}, RNNLM is more likely to pick the tokens with high probabilities during sampling in the generation process. Because of the randomness added by sampling, RNNLM can choose similar alternatives given the same context. Assume we insert tags before the corresponding words (during sentence linearization) to train the RNNLM, when predicting the next token given \emph{``I have booked a flight to"}, the probability of \emph{``S-LOC"} is much higher than the other choices, since the RNNLM has seen many similar examples in training data, such as \emph{``a train to S-LOC"}, \emph{``a trip to S-LOC"} and so on. Then we predict the following word given \emph{``I have booked a flight to S-LOC"}. In the training data, all \emph{``S-LOC"} are followed by location words, so \emph{``London"}, \emph{``Paris"}, \emph{``Tokyo"}, etc., are all possible choices, and their probabilities are very close. Due to the added randomness, the model can choose any one of them. Tokens are predicted in a similar way when we insert tags after the corresponding words, except that words are predicted before the tags.

\subsection{Post-Processing}
\label{sec:post_process}
The generated sequences are in the linearized format, so they need to be converted to the same format as the gold data. We also introduce several straightforward rules to clean the generated data: 1) Delete sentences with no tags; 2) Delete sentences where all words are \emph{[unk]}\footnote{To reduce the size of vocabulary when training language model, the words that only appear once in training data are replaced with the unknown token \emph{[unk]}.}; 3) Delete sentences with incorrect tag prefix orders (e.g., having \emph{E-LOC} before \emph{B-LOC} in NER data); 4) Delete sentences that contain same sequences of words but different tags.

\subsection{Conditional Generation}
\label{subsec:conditional_generation}
We propose a conditional generation method to allow the language model to utilize unlabeled data or knowledge bases when they are available in some low-resource scenarios. For example, it could be expensive to annotate a large amount of e-commerce product titles for NER, but much easier to obtain a knowledge base (i.e., dictionary) of product attributes and unlabeled data. We prepend one of these \textbf{condition tags} $\{ \emph{[labeled]}, \emph{[unlabeled]}, \emph{[KB]} \}$ at the beginning of each sequence to mark their origin, where \emph{KB} means the sequence is labeled by matching a knowledge base against the unlabeled data. See Figure~\ref{fig:conditional_generation} for an example. This allows the language model to learn the shared information among these sequences while being aware of the different origins. When generating synthetic data, each word is conditioned on the given condition tag \emph{[labeled]}, denoted as $c$ (conditioning class). After we feed it into the language model, all the LSTM hidden states $\vh$ in the following generation steps contain information of $c$. In addition, $\vh$ also encodes information of all of the other previous tokens in the sequence. When predicting the next token conditioned on $\vh_{t-1}$, the probability $p_{\theta}(w_t|w_{<t})$ in Eq.~\ref{eq:word_prediction} becomes $p_{\theta}(w_t|w_{<t},c)$.\footnote{Condition tag $c$ is also in $w_{<t}$, since it is a special token added to the beginning of each sentence. We write it explicitly to emphasize the conditional effect.} A similar approach 
is used in CTRL \citep{keskar2019ctrl} to control style, task-specific behavior, etc., during text generation.

\section{Experiments} \label{sec:exp}

In this section, we present our experiments in both supervised and semi-supervised settings. In the supervised settings, only gold data are used for augmentation. In the semi-supervised settings, we also leverage unlabeled data and knowledge bases.

\subsection{Basic Models}
\paragraph{Language Model}
We use the language model described in Section~\ref{subsec:lm_and_gen} for synthetic data generation. We modified the decoder of the LSTM-LM model in \citet{kruengkrai2019better} to implement this language model. We set LSTM hidden state size to 512 and embedding size to 300. We use dropout rate 0.5 for the two dropout layers. All language model are trained using Stochastic gradient descent (SGD) with initial learning rate 1 and batch size 32. Learning rate will be decayed by 0.5 in the next epoch if the perplexity on dev set does not improve. We set the maximum number of epochs to 30 and stop training early if the perplexity on dev set does not improve in 3 consecutive epochs. During synthetic data generation, we use the average length of gold sentences in the training set as our maximum sentence length.

\paragraph{Sequence Tagging Model}
We implement a BiLSTM-CRF model \citep{lample2016neural} with the Flair framework \citep{akbik2019flair} to evaluate our data augmentation method on NER and POS tasks.\footnote{The baseline model provided in the original paper is used for evaluating end to end target based sentiment analysis task.} We use a single-layer BiLSTM with hidden state size 512. Dropout layers are applied before and after the BiLSTM layer with dropout rate 0.5. All sequence tagging models are trained using Adam \citep{kingma2014adam} with initial learning rate 1e-3 and batch size 32. Learning rate is decayed by 0.5 if the performance on dev set does not improve in 3 consecutive epochs. We stop training when the learning rate drops below 1e-5 or number of epochs reaches 100. We use the pre-trained 300-dimensional fastText word embeddings \citep{bojanowski2017enriching} for all languages. 

We employ relatively simple basic models because: 1) They help to avoid the possible overfitting problems due to the small data size under the low resource setting; 2) They allow more faithful understanding on the effects of the proposed data augmentation method.

\label{sec:experiment}
\subsection{Supervised Experiments}
To verify the effectiveness of our data augmentation method in the supervised settings, we evaluate it on three different tagging tasks, including NER, POS and E2E-TBSA. Most of the prior works rely on additional information, so we use random deletion (\textbf{rd}) \citep{wei2019eda} as our baseline, where 5\% of the words\footnote{For NER and E2E-TBSA, the whole entity is deleted if a selected word appears within an entity span.} and the corresponding tags in training data are randomly deleted. See Table~\ref{table:supervised_approaches} for the notations of the methods used in our supervised experiments.

\begin{table}[!ht]
\centering
\scalebox{0.8}{\begin{tabular}{@{}l@{~}@{~}p{8cm}@{}}
\toprule
\textbf{Method} & \textbf{Description}\\
\midrule
\textbf{gold} & Only use the gold data.  \\
\textbf{gen} & Our method. Generate synthetic data with the language models, and oversample gold data. \\
\textbf{rd} & Baseline method. Generate synthetic data by random deletion, and oversample gold data with the same ratio as \textbf{gen}. \\
\textbf{rd\textsuperscript{*}} & Baseline method. Similar to \textbf{rd}, except that gold and synthetic data are equally sampled.  \\
\bottomrule
\end{tabular}}
\caption{Data sources for the supervised setting.}
\label{table:supervised_approaches}
\end{table}

\subsubsection{Named Entity Recognition}

\paragraph{Dataset}
We evaluate our proposed methods on the CoNLL2002/2003 NER data \citep{tjong-kim-sang-2002-introduction,tjong-kim-sang-de-meulder-2003-introduction}, with four languages: English, German, Dutch and Spanish. Besides, we evaluate our methods on Thai and Vietnamese NER data, which are product titles obtained from major e-commerce websites in Southeast Asian countries and annotated with 11 product attribute NER tags, including \emph{PRODUCT}, \emph{BRAND}, \emph{CONSUMER\_GROUP}, \emph{MATERIAL}, \emph{PATTERN}, \emph{COLOR}, \emph{FABRIC}, \emph{OCCASION}, \emph{ORIGIN}, \emph{SEASON} and \emph{STYLE}. See Appendix for the statistics of the Thai and Vietnamese NER data used in our experiments.

\paragraph{Experimental Settings}
In addition to evaluating our method on the full training data, we also randomly sample 1k, 2k, 4k, 6k and 8k sentences for each language to verify its robustness in low-resource settings. We use the original development and test data. The language models are trained on the above sampled sentences, and then used to generate synthetic training data following the steps described in Section~\ref{sec:method}.

For each new batch of 1k sentences generated by the trained language model, we measure the percentage of new 1-gram tokens that appears in previous batches. Once the percentage exceeds 99\%, we will stop data generation. Then we post-process (Section \ref{sec:post_process}) the generated data, and add them to gold training data for tagging model training. For \textbf{rd} and \textbf{gen}, we oversample gold data by repeating them 4 times (shuffled) in the training set. For random deletion, we also report the result when gold and synthetic data are equally sampled, denoted by \textbf{rd\textsuperscript{*}}. Additional details on hyperparameter selection for the oversampling ratios can be found in Appendix A.3. Following \citet{lample2016neural}, the IOBES tagging scheme is used when training the language models and sequence tagging models described above.

\paragraph{Results and Analysis}
We report the average results of 3 runs in Table~\ref{table:conll_ner}. Our method shows consistent performance improvement for all languages. Especially for the smaller sampled sets, our method demonstrates more significant performance improvement. In particular, the proposed method achieved average 1.93 and 1.38 point improvement compared with the baseline methods in the 1k and 2k settings, respectively.

\begin{table}[t!]
\centering
\scalebox{0.75}{\begin{tabular}{@{~}cl@{~}ccccccc@{~}}
\toprule
\textbf{Lang.} & \textbf{Method} & \textbf{1k} & \textbf{2k} & \textbf{4k} & \textbf{6k} & \textbf{8k} & \textbf{all} \\
\midrule
\multirow{4}{*}{\textbf{en}} & gold & 58.06 & 67.85 & 74.55 & 77.16 & 80.30 & 83.04 \\
& ~+rd\textsuperscript{*} & 59.42 & 67.23 & 74.51 & 77.39  & 80.31  & 83.39          \\
& ~+rd & 58.97 & 67.81 & 74.77 & 77.35 & 80.59  & 83.25          \\
& ~+gen & \textbf{61.15} & \textbf{70.61} & \textbf{76.82} & \textbf{79.18} & \textbf{81.02} & \textbf{83.74} \\
\midrule
\multirow{4}{*}{\textbf{de\tablefootnote{German has large out-of-vocabulary rate when using static fastText embedding, which leads to lower F1 score compared with other languages.}}} & gold & 29.71  & \textbf{41.07} & 49.55 & 53.30 & 56.17 & 61.10 \\
& ~+rd\textsuperscript{*} & 29.89  & 40.29  & 49.27  & 52.33  & 55.70 & 60.69 \\
& ~+rd & 30.83 & 40.36   & 49.24   & 53.54  & 55.60   & 60.55          \\
& ~+gen &\textbf{31.83} & 40.92 & \textbf{49.79} & \bf{53.63} & \textbf{56.94} & \textbf{62.44} \\
\midrule
\multirow{4}{*}{\textbf{es}} & gold & 58.14  & 67.42  & 74.21 & 77.44 & 78.90 & 79.27          \\
& ~+rd\textsuperscript{*} & 58.22  & 66.98  & 75.08   & 77.64   & 79.11   & 80.01          \\
& ~+rd & 59.67  & 68.53  & 75.21 & 77.79  & 79.12    & 80.26          \\
& ~+gen & \textbf{61.76} & \textbf{68.62} & \textbf{76.15} & \textbf{78.20} & \textbf{79.83} & \textbf{80.73}  \\
\midrule
\multirow{4}{*}{\textbf{nl}} & gold & 37.04 & 48.61 & 57.78 & 61.08 & 64.59 & 70.89   \\
& ~+rd\textsuperscript{*} & 35.10    & 46.45   & 56.83  & 60.49  & 63.09   & 69.42          \\
& ~+rd & \textbf{39.39} & 48.44   & 59.38  & 61.48 & 64.44   & 70.36          \\
& ~+gen & 38.87  & \textbf{50.41} & \textbf{59.90} & \textbf{63.19} & \textbf{65.82} & \textbf{72.71}  \\
\midrule
\multirow{4}{*}{\textbf{vi}} & gold & 55.98  & 62.42  & 69.01 & 70.75  & 72.12 & 76.14       \\
& ~+rd\textsuperscript{*} & 55.67 & 63.57 & 68.47 & 70.87 & 72.08 & 76.43 \\
& ~+rd & 56.24 & 63.08 & 68.63 & 71.15 & 72.22 & 76.83 \\
& ~+gen & \textbf{60.01} & \textbf{65.43} & \textbf{70.36} & \textbf{72.55} & \textbf{74.11} & \textbf{77.39} \\
\midrule
\multirow{4}{*}{\textbf{th}} & gold & 49.88 & 55.79 & 61.75 & 63.10 & 64.94 & 67.71   \\
& ~+rd\textsuperscript{*} & 50.46 & 56.98 & 62.12 & 64.19 & 66.47 & 67.81 \\
& ~+rd & 50.52 & 57.42 & 61.51 & 64.59 & 66.07 & 67.97 \\
& ~+gen & \textbf{54.02} & \textbf{59.36} & \textbf{63.94} & \textbf{66.21} & \textbf{68.05} & \textbf{69.86} \\
\bottomrule
\end{tabular}}
\caption{Named entity recognition micro F1.}
\label{table:conll_ner}
\end{table}

\paragraph{Tag-Word vs. Word-Tag}
As discussed in Section~\ref{ssec:sentence_linearization}, there are two ways to perform sentence linearization: 1) insert tags before the corresponding words (Tag-Word); 2) insert tags after the corresponding words (Word-Tag). Keeping all of the other settings same, we find Tag-Word outperforms Word-Tag in NER tasks (as shown in Appendix A.2). One possible reason is that, Tag-Word is more consistent with the Modifier-Noun pattern, which appears more often in the training data during language modeling. Therefore, we use Tag-Word for all the NER experiments.

\subsubsection{Part of Speech Tagging}
\paragraph{Dataset}
We use the POS data from Universal Dependencies treebanks\footnote{https://universaldependencies.org/} for evaluation on this task. We evaluate on five languages, including English, Spanish, Czech, Romanian and Japanese. Each language has multiple corpora in the Universal Dependencies treebanks, so we merge the corpora to build one dataset for three languages: English, Spanish and Czech. For English, we merge \emph{GUM}, \emph{ParTUT}, \emph{PUD} and \emph{Lines}. For Spanish, we merge \emph{AnCora} and \emph{GSD}. For Czech, we merge \emph{PDT}, \emph{FicTree}, \emph{CLTT} and \emph{CAC}. We have also evaluated our model on Japanese (\emph{GSD}) and Romanian (\emph{RRT}), which are either spoken by a much smaller population or from a different language family.

\paragraph{Settings and Results}
We follow similar experimental settings as NER task. The same language model and BiLSTM-CRF sequence tagging model are used for synthetic data generation and POS tagging respectively. Different from NER, Word-Tag shows slightly better performance in POS tasks (refer to Appendix A.2).
We present the average Word-Tag results of 3 runs in Table~\ref{table:ud_pos}. Our method demonstrates consistent performance improvement for all languages. Similar to NER, our method demonstrates more significant performance improvement for smaller sampled sets on POS tagging. In particular, the proposed method achieved average 0.56, 0.60 and 0.46 point improvement compared with the baseline methods in the 1k, 2k and 4k settings, respectively.

\begin{table}[t!]
\centering
\scalebox{0.75}{\begin{tabular}{@{~}cl@{~}cccccc@{~}}
\toprule
\textbf{Lang.} & \textbf{Method} & \textbf{1k} & \textbf{2k} & \textbf{4k} & \textbf{6k} & \textbf{8k} & \textbf{Full} \\
\midrule
\multirow{4}{*}{\textbf{en}} & gold & 79.18 & 82.17 & 85.83 & 88.62 & 90.21  & 93.00\\
& ~+rd\textsuperscript{*} & 79.28 & 82.42 & 85.82 & 88.55 & 90.07 & 92.89 \\
& ~+rd & 79.38 & 82.50 & 86.08 & 88.80 & 90.15 & 92.96 \\
& ~+gen & \bf{79.76} & \bf{82.90} & \bf{86.31} & \bf{88.99} & \bf{90.56} & \bf{93.29} \\
\midrule
\multirow{4}{*}{\textbf{es}} & gold & 88.28 & 90.79 & 92.82 & 93.80 & 94.43  & 96.40 \\
& ~+rd\textsuperscript{*} & 88.25 & 90.94 & 92.84 & 93.76 & 94.48 & 96.41\\
& ~+rd & 88.17 & 90.78 & 92.79 & 93.67 & 94.28 & \bf{96.45}\\
& ~+gen & \bf{88.77} & \bf{91.04} & \bf{93.12} &	\bf{93.93} & \bf{94.64} & \bf{96.45}\\
\midrule
\multirow{3}{*}{\textbf{cz}} & gold & 80.10 & 84.46 & 88.88 & 90.67 & 92.03 & 97.52\\
& ~+rd\textsuperscript{*} & 79.83 & 84.29 & 88.64 & 90.43 & 91.95 & 97.57 \\
& ~+rd & 80.11 & 84.50 & 88.99 & 90.66 & 91.86 & 97.60 \\
& ~+gen & \bf{80.65} & \bf{85.17} & \bf{89.58} & \bf{91.22} & \bf{92.49} & \bf{97.63} \\
\midrule
\multirow{3}{*}{\textbf{ro\tablefootnote{UD-RRT full train set has 8k sentences for Romanian.}}} & gold & 86.69 & 89.57 & 92.73 & 93.84 & 94.54 & 94.54 \\
& ~+rd\textsuperscript{*} & 86.42 &	89.58 &	92.50 &	93.89 &	94.64 & 94.64 \\
& ~+rd & 86.62 & 89.46 & 92.55 & 93.84 & 94.73 & 94.73  \\
& ~+gen & \bf{87.29} &	\bf{90.66} & \bf{93.44} & \bf{94.61} & \bf{95.17} & \bf{95.17}\\
\midrule
\multirow{3}{*}{\textbf{ja\tablefootnote{UD-GSD full train set has 7k sentences for Japanese.}}} & gold & 90.19 & 91.44 & 93.59 & 94.41 & - & 95.08
\\
& ~+rd\textsuperscript{*} & 90.00 &	91.41 &	93.66 &	94.62 & - &	94.93 \\
& ~+rd & 89.53 & 91.76 & 93.62 & 94.59 & - 	& 95.18 \\
& ~+gen & \bf{91.00} & \bf{92.51} &	\bf{94.12} & \bf{95.21} & - & \bf{95.45} \\
\bottomrule
\end{tabular}}
\caption{POS tagging accuracy.}
\label{table:ud_pos}
\end{table}

\subsubsection{Target Based Sentiment Analysis}
\paragraph{Dataset}
We use the laptop and restaurant review datasets processed by \citet{li2019unified} for evaluation on E2E-TBSA, which was initially obtained from SemEval ABSA challenges \citep{pontiki-etal-2014-semeval,pontiki-etal-2015-semeval,pontiki-etal-2016-semeval}. We merge these two review datasets, regard $10\%$ randomly held-out training data as the dev set, and randomly sample smaller sets from the remaining training data for low-resource settings. The original test sets are merged as our test set. 

\paragraph{Settings and Results}
We follow similar experimental settings as NER and POS tasks, except that the same sequence tagging model released by \citet{li2019unified} is used for evaluation. Here  Tag-Word shows better results (refer to Appendix A.2), plausibly it is because the unified tags (e.g. \emph{B-POS}, and \emph{B-NEG}) are similar to noun modifiers and Tag-Word is more consistent with the Modifier-Noun pattern. We present the average Tag-Word results of 3 runs in Table~\ref{table:e2etbsa}. Our method demonstrates performance improvement for 4k and above. Compared with the NER and POS datasets, the E2E-TBSA dataset has much fewer labels, so the results are less stable.

\begin{table}[th!]
\centering
\scalebox{0.78}{\begin{tabular}{lcccc}
\toprule
\textbf{Method} & \textbf{2k} & \textbf{4k} & \textbf{all(6k)} \\
\midrule
gold & 56.31 & 60.43 & 63.18  \\
~+rd\textsuperscript{*} & \bf{57.92} & 61.75 & 63.66 \\
~+gen & 57.07 & \bf{62.66} & \bf{65.86}  \\
\bottomrule
\end{tabular}}
\caption{ E2E-TBSA micro F1.}
\label{table:e2etbsa}
\end{table}

\subsection{Semi-supervised Experiments}
In this section we evaluate the effectiveness of our method in two semi-supervised settings: 1) only unlabeled data are available; 2) both unlabeled data and knowledge base are available. See Table~\ref{table:semisupervised_approaches} for the notations of the methods used in our semi-supervised experiments.

\begin{table}[htb!]
\centering
\scalebox{0.8}{\begin{tabular}{@{}l@{~}@{~}p{8cm}@{}}
\toprule
\textbf{Method} & \textbf{Description}\\
\midrule
\textbf{gold} & Supervised method. Only use the gold data.  \\
\textbf{wt} & Baseline method. Annotate unlabeled data with a weak tagger (i.e. a tagging model trained on the gold data).  \\
\textbf{gen\textsubscript{ud}} & Our method. Generate synthetic data with LM, where LM is trained on gold data and unlabeled data. \\
\textbf{kb} & Baseline method. Annotate unlabeled data with knowledge base.  \\
\textbf{gen\textsubscript{kb}} & Our method. Generate synthetic data with LM, where LM is trained on gold data and knowledge base annotated data. \\
\bottomrule
\end{tabular}}
\caption{Data sources for the semi-supervised setting.}
\label{table:semisupervised_approaches}
\end{table}

\subsubsection{Only Using Unlabeled Data}
\label{sssec:semi_ner_unlabeled}

\paragraph{Dataset}
We use CoNLL2003 English NER data \citep{tjong-kim-sang-de-meulder-2003-introduction} for evaluation. In addition to the gold NER training data, we utilize unlabeled data for semi-supervised training. The Stanford CoreNLP tokenizer \citep{manning-EtAl:2014:P14-5} is used to tokenize Wikipedia sentences. 

\paragraph{Experimental Settings}
Similar to the above experiments, we use 1k, 2k, 4k, 6k and 8k sentences randomly sampled from NER gold data as well as the full dataset to evaluate our method. For fair comparison, we only use the same set of 10k sentences randomly sampled from Wikipedia dump in both of our and baseline methods. Let $D_{gold}$ and $D_{unlabeled}$ be the sampled gold NER data and the Wikipedia data, respectively. 

In our method, $D_{gold}$ and $D_{unlabeled}$ are concatenated to train language models, following the steps described in Section~\ref{subsec:conditional_generation}. 
Then we use the language models to generate synthetic data, from which 20k randomly sampled sentences are combined with $D_{gold}$ to train NER models. We use \textbf{gen\textsubscript{ud}} to denote this data generation method. The method that employs weak taggers to annotate $D_{unlabeled}$ is used as our baseline, denoted by \textbf{wt}. The weak taggers in this experiment are the NER models trained on $D_{gold}$. We use the same NER model (BiLSTM-CRF) and hyperparameters to evaluate our and baseline methods. When training the language model, we equally sample sentences from $D_{gold}$ and $D_{unlabeled}$. When training the NER models, we oversample the gold data by repeating $D_{gold}$ 4 times to create a shuffled training file.

\paragraph{Results and Analysis}
We report F1 of \textbf{wt} and \textbf{gen\textsubscript{ud}} (average of 3 runs) in Table~\ref{table:semi_ner_unlabeled}. Our method outperforms the baseline method \textbf{wt} on all settings. Moreover, it  would be a promising direction to further explore our model's capability by using a larger amount of unlabeled data. It is not very convenient to utilize a large amount of unlabeled data in the baseline method \textbf{wt}, since this will directly increase the amount of augmented data. As a result, some of augmented data may not be utilized before sequence tagging models converge. However, our method \textbf{gen\textsubscript{ud}} can conveniently utilize a large amount of unlabeled data to train the language models, thereby improving data augmentation quality directly. When the amount of unlabeled data is much larger, we can pretrain language models with the unlabeled data, and then finetune them with labeled data.

\begin{table}[t!]
\centering
\scalebox{0.82}{\begin{tabular}{lcccccc}
\toprule
\textbf{Method} & \textbf{1k} & \textbf{2k} & \textbf{4k} & \textbf{6k} & \textbf{8k} & \textbf{all} \\
\midrule
gold & 58.06 & 67.85 & 74.55 & 77.16 & 80.30 & 83.04 \\
\midrule
\rowcolor{lightyellow}  ~+wt & 65.12 & 72.43 & 77.90 & 79.41 & 81.36 & 84.00 \\
\rowcolor{lightyellow}  ~+gen\textsubscript{ud} & \bf{66.19} & \bf{73.00} & \bf{78.08} & \bf{79.75} & \bf{81.98} & \bf{84.33} \\
\midrule
\rowcolor{lavender}~+kb & \bf{67.36} & 72.86 & 77.15 & 79.33 & 81.91 & 83.69 \\
\rowcolor{lavender}~+gen\textsubscript{kb} & 66.67 & \bf{73.54} & \bf{78.32} & \bf{79.98} & \bf{81.93} & \bf{84.03} \\
\bottomrule
\end{tabular}}
\caption{Semi-supervised NER F1.}
\label{table:semi_ner_unlabeled}
\end{table}

\subsubsection{Using Unlabeled Data and Knowledge Base}
\paragraph{Dataset}
In addition to the gold training data and unlabeled sentences used in Section~\ref{sssec:semi_ner_unlabeled}, we also try to leverage knowledge base for further performance improvement in this experiment. We build the knowledge base by extracting entities (case sensitive and appearing at least twice) and the corresponding tags from the full gold training data. Besides, we add more \emph{LOC} entities to this knowledge base by including the cities and countries extracted from geonames\footnote{https://datahub.io/core/world-cities}.

\paragraph{Experimental Settings}
We randomly sample the gold NER data and the Wikipedia data in the same way as Section~\ref{sssec:semi_ner_unlabeled}, where the sampled sentences are denoted by $D_{gold}$ and $D_{unlabeled}$, respectively. Our knowledge base is used to annotate $D_{unlabeled}$ by finding longest forward matches (from left to right) in each sentence. We denote this annotation method by \textbf{kb} and the annotated data by $D_{kb}$. 

In our method, $D_{gold}$ and $D_{kb}$ are concatenated to train language models, following the steps described in Section~\ref{subsec:conditional_generation}. 
Then we use the language models to generate synthetic data, from which 20k randomly sampled sentences are combined with $D_{gold}$ to train the NER models. 
We use \textbf{gen\textsubscript{kb}} to denote this data generation method and compare it with the baseline method \textbf{kb}. Similar to the above experiments, we oversample $D_{gold}$ when training the language and NER models.

\paragraph{Results and Analysis}
We present F1 of \textbf{kb} and \textbf{gen\textsubscript{kb}} (average of 3 runs) in Table~\ref{table:semi_ner_unlabeled}. The baseline method \textbf{kb} exhibits very strong performance when the size of $D_{gold}$ is small, since we use a large knowledge base of countries and cities for annotation, and location names are less ambiguous compared with the other types of entities. However, our method still outperforms \textbf{kb} when the size of $D_{gold}$ is larger than 2k, which shows that our method is more robust to the noises in $D_{kb}$ when a slightly larger amount of gold data are available.

\section{A Closer Look at Synthetic Data}
\label{sec:gen_data}

In this section, we explore in more details why the synthetic data generated by our method can help improve sequence tagging performance. Through a closer look at the generated data, we have several interesting findings. 

\begin{figure}[t!]
\centering
\includegraphics[scale=0.4]{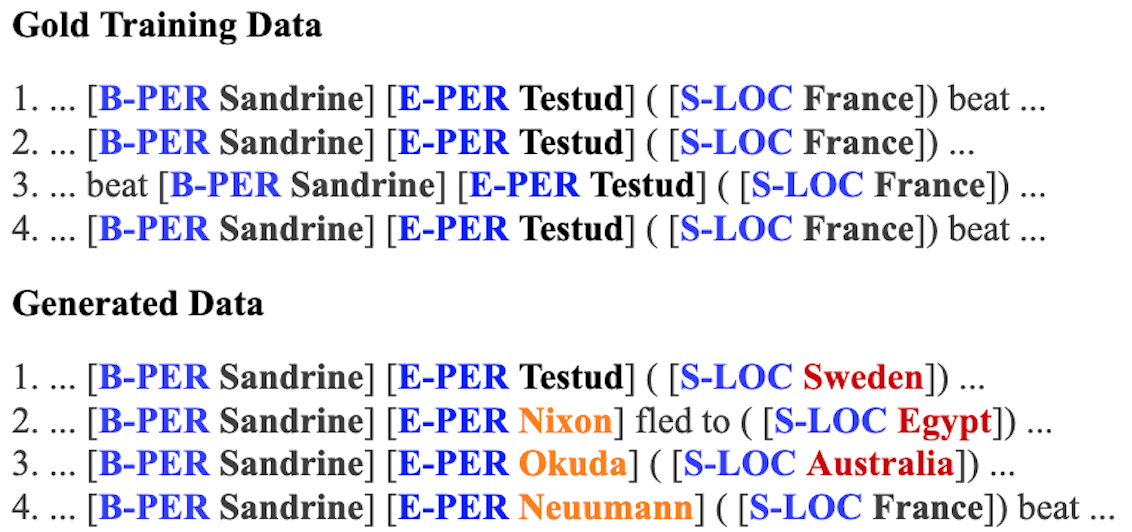}
\caption{An illustration of diversity of generated data. The name ``\emph{Sandrine}" in the gold training data always pairs up with ``\emph{Testud}" in sentences.}
\label{fig:analysis_diversity}
\end{figure}

\paragraph{More Diversity}
The generated synthetic data introduces more diversity to help reduce overfitting. As the example shown in Figure~\ref{fig:analysis_diversity}, the name ``\emph{Sandrine}" in the gold training data always pairs up with ``\emph{Testud}" in different sentences. However, in the generated data, we can see new names have been generated like ``\emph{Sandrine Nixon}", ``\emph{Sandrine Okuda}" and ``\emph{Sandrine Neuumann}". Meanwhile, the locations in  the sentences have been replaced with new countries like ``\emph{Sweden}", ``\emph{Egypt}" and ``\emph{Australia}". With these synthetic data, the model can focus on learning the pattern of contexts that entities appear in, instead of simply memorizing ``\emph{Sandrine Testud}" as a person name and ``\emph{France}" as a location. 

\begin{figure}[t!]
\centering
\includegraphics[scale=0.52]{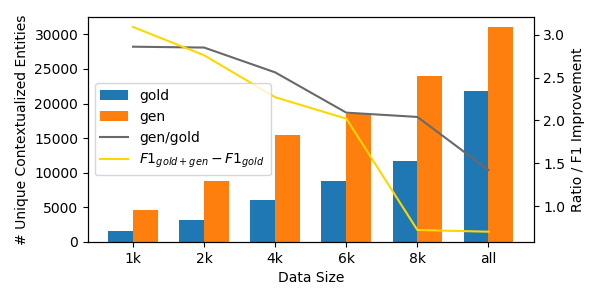}
\caption{Statistics of unique contextualized entities.}
\label{fig:contextualized_entity}
\end{figure}

To quantitatively measure the diversity brought in by our method and its impact, we have done some statistical analysis of the contextualized entities (CEs) in the generated data of the supervised English NER. A CE refers to the combination of an entity and its 1-gram contexts. For example, in the sentence ``\emph{The [B-ORG European] [E-ORG Commission] said ...}", the entity is ``\emph{European Commission}" and its CE is ``\emph{The European Commission said}". 
As the shown in Figure~\ref{fig:contextualized_entity}, we calculate the number of unique CEs in the gold training data, the number of new unique CEs in the generated data, and the ratio of the two numbers. We also plot here the F1 improvement of our method (i.e. gold+gen) over only using the gold data, as given in Table \ref{table:conll_ner}. 
We can see that our method generates a large number of new CEs, and such diversity strengthens the robustness of the trained model.  
When the ratio is higher, the F1 improvement is more significant, which also shows that our method does help ease the low-resource problem by generating rich new entities and contexts.
Refer to the Appendix A.5 for statistics of unique entities (without context), which shows the same conclusion.

\paragraph{Efficient Utilization of Unlabeled Data}
When unlabeled data are available, our method is flexible to utilize them for semi-supervised training. We find many interesting examples in the synthetic data that show our method can effectively use the unlabeled data to extract useful information. In an example generated by our method ``\emph{... the [B-ORG Bank] [I-ORG of] [E-ORG Alabama] ...}", the word ``\emph{Alabama}” has never appeared in gold NER training data. However, our language model learned that ``\emph{Alabama}” (from unlabeled data) is very similar to the other location words that appear in both gold training data and unlabelled data. So when generating the synthetic data, the language model can use this word in a similar context, or even create new entities (``\emph{Bank of Alabama}" in this example has never appeared in gold or unlabeled data).

\section{Conclusion}
\label{sec:conclusion}

In this paper, we show that language models can be used to generate high quality synthetic data for sequence tagging tasks. The generated data introduce more diversity to reduce overfitting, since they are generated from scratch instead of modifying the gold training data. Our proposed method demonstrates promising performance improvements on various tagging tasks, especially in the low-resource settings. Besides, experiments demonstrate that our method can also effectively utilize unlabeled data and knowledge base for semi-supervised training.

\section*{Acknowledgements}
\label{sec:acknowledgements}

This research is partly supported by the Alibaba-NTU Singapore Joint Research Institute, Nanyang Technological University. Linlin Liu would like to thank the support from Interdisciplinary Graduate School, Nanyang Technological University.

\bibliographystyle{acl_natbib}
\bibliography{main}

\appendix

\section{Appendix}

\subsection{Statistics of Thai and Vietnamese NER Data}
We present the number of sentences in Thai and Vietnamese NER data in Table~\ref{table:th_vi_stats}.

\begin{table}[h!]
\centering
\scalebox{0.7}{\begin{tabular}{cccc}
\toprule
\textbf{Lang.} & \textbf{train} & \textbf{dev} & \textbf{test} \\
\midrule
\bf{vi} & 18,922 & 500 & 500 \\
\bf{th} & 11,272 & 499 & 490 \\
\bottomrule
\end{tabular}}
\caption{Number of sentences in TH and VI NER data.}
\label{table:th_vi_stats}
\end{table}



\subsection{Experiments on Tag-Word vs. Word-Tag}
We conduct experiments to compare the performance of Tag-Word and Word-Tag for the tagging tasks. All of the other settings are same as the corresponding experiments presented in the main paper. Results are reported in Table~\ref{table:tag_pos_ner} to \ref{table:tag_pos_tbsa}. Tag-Word yields better average performance for NER and E2E-TBSA, while Word-Tag slightly outperforms Tag-Word for POS tagging.

\begin{table}[h!]
\centering
\scalebox{0.6}{\begin{tabular}{clccccccc}
\toprule
\textbf{Lang.} & \textbf{Method} & \textbf{1k} & \textbf{2k} & \textbf{4k} & \textbf{6k} & \textbf{8k} & \textbf{full} & \textbf{average}  \\
\midrule
\multirow{2}{*}{\textbf{en}} & Tag-Word &  \bf{59.39} & \bf{69.48} & \bf{75.68} & \bf{78.65} & \bf{80.19} & \bf{83.70} & \bf{74.52} \\
& Word-Tag & 58.97 & 67.32 & 75.45 & 78.06 & 80.43 & 83.58 & 73.97 \\
\bottomrule
\end{tabular}}
\caption{CoNLL NER F1: Tag-Word vs. Word-Tag.}
\label{table:tag_pos_ner}
\end{table}

\begin{table}[h!]
\centering
\scalebox{0.7}{\begin{tabular}{clcccccc}
\toprule
\textbf{Lang.} & \textbf{Method} & \textbf{1k} & \textbf{2k} & \textbf{4k} & \textbf{8k} & \textbf{15k} & \textbf{average}  \\
\midrule
\multirow{2}{*}{\textbf{en}} & Tag-Word & 79.06 & 82.43 &	85.93 & \bf{90.38} &	\bf{92.75} & 86.11  \\
& Word-Tag & \bf{79.18} & \bf{82.64} & \bf{86.13} & 90.33 & 92.68 & \bf{86.19} \\
\bottomrule
\end{tabular}}
\caption{Universal Dependencies POS accuracy: Tag-Word vs. Word Tag.}
\label{table:tag_pos_pos}
\end{table}

\begin{table}[h!]
\centering
\scalebox{0.7}{\begin{tabular}{clcccc}
\toprule
\textbf{Lang.} & \textbf{Method} & \textbf{2k} & \textbf{4k} & \textbf{full(6k)} & \textbf{average}  \\
\midrule
\multirow{2}{*}{\textbf{en}} & Tag-Word & 54.22 & \bf{61.72} & \bf{62.88} & \bf{59.61}  \\
& Word-Tag & \bf{55.58} & 59.42 & 61.65 & 58.88 \\
\bottomrule
\end{tabular}}
\caption{E2E-TBSA micro F1: Tag-Word vs. Word-Tag.}
\label{table:tag_pos_tbsa}
\end{table}

\subsection{Experiments on Oversampling Ratios}
We conduct experiments to compare different oversampling ratios for NER task. Results are reported in Table~\ref{table:os_ratio}. The notation gold$\times N$ means we oversample gold by repeating it $N$ times in the shuffled static training data.

\begin{table}[t!]
\centering
\scalebox{0.7}{\begin{tabular}{clcccccc}
\toprule
\textbf{Lang.} & \textbf{Method} & \textbf{1k} & \textbf{2k} & \textbf{4k} & \textbf{average}  \\
\midrule
\multirow{5}{*}{\textbf{en}} & gold$\times1$ & 59.74 & 69.14 & 76.48 & 68.45 \\
& gold$\times2$ & 60.92 & 69.79 & 76.57 & 69.09 \\
& gold$\times3$ & 61.13 & 70.42 & 74.92 & 67.24 \\
& gold$\times4$ & 61.15 & \bf{70.61} & \bf{76.82} & \bf{69.53} \\
& gold$\times5$ & \bf{61.43} & 70.38 & 76.43 & 69.41 \\
\bottomrule
\end{tabular}}
\caption{CoNLL NER F1: comparison on different oversampling ratios.}
\label{table:os_ratio}
\end{table}

\subsection{Semi-supervised Experiments on Part of Speech Tagging}

\paragraph{Dataset}
We use the English POS data from Universal Dependencies treebanks for evaluation on this task. We merge \emph{GUM}, \emph{ParTUT}, \emph{PUD} and \emph{Lines} corpora to build the English dataset. Similar to the semi-supervised experiments on NER, we also utilize unlabeled Wikipedia sentences for training.

\paragraph{Experimental Settings}
We use 1k, 2k, 4k, 6k and 8k sentences randomly sampled from English POS gold data as well as the full dataset to evaluate our method. We follow the same experimental setting as the semi-supervised experiments on NER to generate synthetic data, train the sequence tagging models and evaluate on the POS test data.

\paragraph{Results and Analysis}
We report accuracy of \textbf{wt} and \textbf{gen\textsubscript{ud}} (average of 3 runs) in Table~\ref{table:semi_pos_unlabeled}. Our method outperforms the baseline method \textbf{wt} when the number of gold sentences are less than 8k. When the number of gold sentences are more than 8k, the performance of our method is comparable with \textbf{wt}.

\begin{table}[hbt!]
\centering
\scalebox{0.7}{\begin{tabular}{lcccccc}
\toprule
\textbf{Method} & \textbf{1k} & \textbf{2k} & \textbf{4k} & \textbf{6k} & \textbf{8k} & \textbf{all} \\
\midrule
gold & 79.18 & 82.17 & 85.83 & 88.62 & 90.21  & 93.00 \\
\midrule
~+wt & 81.11 & 84.00 & 86.91 & 89.64 & \bf{90.88} & \bf{93.20} \\
~+gen\textsubscript{ud} & \bf{82.11} & \bf{84.93} & \bf{87.52} & \bf{89.98} & 90.84 & 93.12 \\

\bottomrule
\end{tabular}}
\caption{Semi-supervised POS accuracy.}
\label{table:semi_pos_unlabeled}
\end{table}

\subsection{Synthetic Data Diversity: Unique Entities}
\begin{figure}[h!]
\centering
\includegraphics[scale=0.5]{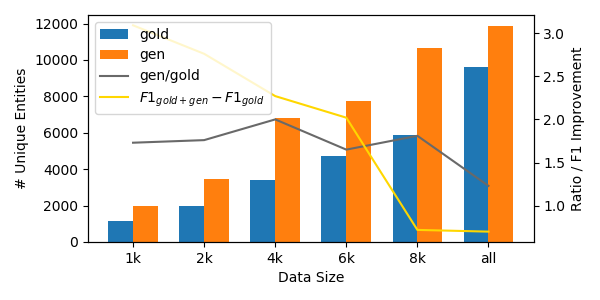}
\caption{Statistics of unique entities (without context)}
\label{fig:unique_entity}

\end{figure}

To quantitatively measure the diversity introduced by our method in the supervised English NER tasks, we count the number of unique entities (without context) in the gold and generated data. Results are presented in Figure~\ref{fig:unique_entity}.

\subsection{Average Runtime}
Table~\ref{table:running_time} is an illustration of the average runtime of our models in English NER, POS and E2E-TBSA tasks and RNNLM.

\begin{table}[h!]
\centering
\scalebox{0.7}{\begin{tabular}{lcccccc}
\toprule
\textbf{Task} & \textbf{1k} & \textbf{2k} & \textbf{4k} & \textbf{6k} & \textbf{8k} & \textbf{all}  \\
\midrule
NER & 26.5 & 70.5 & 124.4 & 167.9 & 216.3 & 393.2 \\
POS & 83.6 & 112.3 & 231.1 & 257.7 & 277.2 & 298.0 \\
E2E-TBSA & - & 89.3 & 150.8 & 269.2 & - & - \\
RNNLM  & 0.7 & 1.1 & 2.0 & 2.7 & 3.3 & 3.9 \\
\bottomrule
\end{tabular}}
\caption{Average runtime (min).}
\label{table:running_time}
\end{table}

\subsection{Computing Infrastructure}
We conduct our experiments on NVIDIA V100 GPU.

\end{document}

%% file: math_commands.tex
\usepackage{amsmath}
\usepackage{amsfonts,bm}

\def\eqref#1{equation~\ref{#1}}

\def\1{\bm{1}}

\def\vd{{\bm{d}}}

\def\vh{{\bm{h}}}

\def\vs{{\bm{s}}}

\def\vx{{\bm{x}}}

\def\m1{{\bm{1}}}

\def\mM{{\bm{M}}}

\DeclareMathAlphabet{\mathsfit}{\encodingdefault}{\sfdefault}{m}{sl}
\SetMathAlphabet{\mathsfit}{bold}{\encodingdefault}{\sfdefault}{bx}{n}

\newcommand{\dropout}{\mathrm{dropout}}
\newcommand{\LSTM}{\mathrm{LSTM}}



%% file: main.bbl
\begin{thebibliography}{50}
\expandafter\ifx\csname natexlab\endcsname\relax\def\natexlab#1{#1}\fi

\bibitem[{Akbik et~al.(2019)Akbik, Bergmann, Blythe, Rasul, Schweter, and
  Vollgraf}]{akbik2019flair}
Alan Akbik, Tanja Bergmann, Duncan Blythe, Kashif Rasul, Stefan Schweter, and
  Roland Vollgraf. 2019.
\newblock Flair: An easy-to-use framework for state-of-the-art nlp.
\newblock In \emph{Proceedings of the 2019 Conference of the North American
  Chapter of the Association for Computational Linguistics (Demonstrations)},
  pages 54--59.

\bibitem[{Akbik et~al.(2018)Akbik, Blythe, and Vollgraf}]{akbik2018contextual}
Alan Akbik, Duncan Blythe, and Roland Vollgraf. 2018.
\newblock Contextual string embeddings for sequence labeling.
\newblock In \emph{Proceedings of the 27th International Conference on
  Computational Linguistics}, pages 1638--1649.

\bibitem[{Anaby-Tavor et~al.(2020)Anaby-Tavor, Carmeli, Goldbraich, Kantor,
  Kour, Shlomov, Tepper, and Zwerdling}]{anaby2020not}
Ateret Anaby-Tavor, Boaz Carmeli, Esther Goldbraich, Amir Kantor, George Kour,
  Segev Shlomov, Naama Tepper, and Naama Zwerdling. 2020.
\newblock Do not have enough data? deep learning to the rescue!
\newblock In \emph{Thirty-Third AAAI Conference on Artificial Intelligence}.

\bibitem[{Bari et~al.(2020)Bari, Mohiuddin, and Joty}]{bari2020multimix}
M~Saiful Bari, Tasnim Mohiuddin, and Shafiq Joty. 2020.
\newblock \href {http://arxiv.org/abs/2004.13240} {Multimix: A robust data
  augmentation framework for cross-lingual nlp}.

\bibitem[{Bohnet et~al.(2018)Bohnet, McDonald, Sim{\~o}es, Andor, Pitler, and
  Maynez}]{bohnet2018morphosyntactic}
Bernd Bohnet, Ryan McDonald, Gon{\c{c}}alo Sim{\~o}es, Daniel Andor, Emily
  Pitler, and Joshua Maynez. 2018.
\newblock Morphosyntactic tagging with a meta-bilstm model over context
  sensitive token encodings.
\newblock In \emph{Proceedings of the 56th Annual Meeting of the Association
  for Computational Linguistics (Volume 1: Long Papers)}, pages 2642--2652.

\bibitem[{Bojanowski et~al.(2017)Bojanowski, Grave, Joulin, and
  Mikolov}]{bojanowski2017enriching}
Piotr Bojanowski, Edouard Grave, Armand Joulin, and Tomas Mikolov. 2017.
\newblock Enriching word vectors with subword information.
\newblock \emph{Transactions of the Association for Computational Linguistics},
  5:135--146.

\bibitem[{Chen et~al.(2017)Chen, Sun, Bing, and
  Yang}]{chen-EtAl:2017:EMNLP20171}
Peng Chen, Zhongqian Sun, Lidong Bing, and Wei Yang. 2017.
\newblock Recurrent attention network on memory for aspect sentiment analysis.
\newblock In \emph{Proceedings of EMNLP}, pages 463--472.

\bibitem[{Dong et~al.(2017)Dong, Mallinson, Reddy, and
  Lapata}]{dong-etal-2017-learning}
Li~Dong, Jonathan Mallinson, Siva Reddy, and Mirella Lapata. 2017.
\newblock \href {https://doi.org/10.18653/v1/D17-1091} {Learning to paraphrase
  for question answering}.
\newblock In \emph{Proceedings of the 2017 Conference on Empirical Methods in
  Natural Language Processing}, pages 875--886, Copenhagen, Denmark.
  Association for Computational Linguistics.

\bibitem[{D’Innocente et~al.(2017)D’Innocente, Carlucci, Colosi, and
  Caputo}]{d2017bridging}
Antonio D’Innocente, Fabio~Maria Carlucci, Mirco Colosi, and Barbara Caputo.
  2017.
\newblock Bridging between computer and robot vision through data augmentation:
  a case study on object recognition.
\newblock In \emph{International Conference on Computer Vision Systems}, pages
  384--393. Springer.

\bibitem[{Fadaee et~al.(2017)Fadaee, Bisazza, and Monz}]{fadaee2017data}
Marzieh Fadaee, Arianna Bisazza, and Christof Monz. 2017.
\newblock Data augmentation for low-resource neural machine translation.
\newblock In \emph{Proceedings of the 55th Annual Meeting of the Association
  for Computational Linguistics (Volume 2: Short Papers)}, pages 567--573.

\bibitem[{Fawzi et~al.(2016)Fawzi, Samulowitz, Turaga, and
  Frossard}]{fawzi2016adaptive}
Alhussein Fawzi, Horst Samulowitz, Deepak Turaga, and Pascal Frossard. 2016.
\newblock Adaptive data augmentation for image classification.
\newblock In \emph{2016 IEEE International Conference on Image Processing
  (ICIP)}, pages 3688--3692. Ieee.

\bibitem[{Heinzerling and Strube(2019)}]{heinzerling2019sequence}
Benjamin Heinzerling and Michael Strube. 2019.
\newblock Sequence tagging with contextual and non-contextual subword
  representations: A multilingual evaluation.
\newblock In \emph{Proceedings of the 57th Annual Meeting of the Association
  for Computational Linguistics}, pages 273--291.

\bibitem[{Keskar et~al.(2019)Keskar, McCann, Varshney, Xiong, and
  Socher}]{keskar2019ctrl}
Nitish~Shirish Keskar, Bryan McCann, Lav~R Varshney, Caiming Xiong, and Richard
  Socher. 2019.
\newblock Ctrl: A conditional transformer language model for controllable
  generation.
\newblock \emph{arXiv preprint arXiv:1909.05858}.

\bibitem[{Kingma and Ba(2014)}]{kingma2014adam}
Diederik~P Kingma and Jimmy Ba. 2014.
\newblock Adam: A method for stochastic optimization.
\newblock \emph{arXiv preprint arXiv:1412.6980}.

\bibitem[{Ko et~al.(2017)Ko, Peddinti, Povey, Seltzer, and
  Khudanpur}]{ko2017study}
Tom Ko, Vijayaditya Peddinti, Daniel Povey, Michael~L Seltzer, and Sanjeev
  Khudanpur. 2017.
\newblock A study on data augmentation of reverberant speech for robust speech
  recognition.
\newblock In \emph{2017 IEEE International Conference on Acoustics, Speech and
  Signal Processing (ICASSP)}, pages 5220--5224. IEEE.

\bibitem[{Kobayashi(2018)}]{kobayashi2018contextual}
Sosuke Kobayashi. 2018.
\newblock Contextual augmentation: Data augmentation by words with paradigmatic
  relations.
\newblock In \emph{Proceedings of the 2018 Conference of the North American
  Chapter of the Association for Computational Linguistics: Human Language
  Technologies, Volume 2 (Short Papers)}, pages 452--457.

\bibitem[{Kruengkrai(2019)}]{kruengkrai2019better}
Canasai Kruengkrai. 2019.
\newblock Better exploiting latent variables in text modeling.
\newblock In \emph{Proceedings of the 57th Annual Meeting of the Association
  for Computational Linguistics}, pages 5527--5532.

\bibitem[{Kumar et~al.(2020)Kumar, Choudhary, and Cho}]{kumar2020data}
Varun Kumar, Ashutosh Choudhary, and Eunah Cho. 2020.
\newblock Data augmentation using pre-trained transformer models.
\newblock \emph{arXiv preprint arXiv:2003.02245}.

\bibitem[{Lample et~al.(2016)Lample, Ballesteros, Subramanian, Kawakami, and
  Dyer}]{lample2016neural}
Guillaume Lample, Miguel Ballesteros, Sandeep Subramanian, Kazuya Kawakami, and
  Chris Dyer. 2016.
\newblock Neural architectures for named entity recognition.
\newblock \emph{arXiv preprint arXiv:1603.01360}.

\bibitem[{Li et~al.(2020)Li, Sun, Han, and Li}]{li2020survey}
Jing Li, Aixin Sun, Jianglei Han, and Chenliang Li. 2020.
\newblock A survey on deep learning for named entity recognition.
\newblock \emph{IEEE Transactions on Knowledge and Data Engineering}.

\bibitem[{Li et~al.(2019{\natexlab{a}})Li, Bing, Li, and Lam}]{li2019unified}
Xin Li, Lidong Bing, Piji Li, and Wai Lam. 2019{\natexlab{a}}.
\newblock A unified model for opinion target extraction and target sentiment
  prediction.
\newblock In \emph{Proceedings of the AAAI Conference on Artificial
  Intelligence}, volume~33, pages 6714--6721.

\bibitem[{Li et~al.(2018)Li, Bing, Li, Lam, and Yang}]{li2018aspect}
Xin Li, Lidong Bing, Piji Li, Wai Lam, and Zhimou Yang. 2018.
\newblock \href {https://www.ijcai.org/proceedings/2018/0583.pdf} {Aspect term
  extraction with history attention and selective transformation}.
\newblock In \emph{Proceedings of the Twenty-Seventh International Joint
  Conference on Artificial Intelligence (IJCAI)}, pages 4194--4200.

\bibitem[{Li et~al.(2019{\natexlab{b}})Li, Bing, Zhang, and
  Lam}]{li2019bert-e2e-absa}
Xin Li, Lidong Bing, Wenxuan Zhang, and Wai Lam. 2019{\natexlab{b}}.
\newblock Exploiting {BERT} for end-to-end aspect-based sentiment analysis.
\newblock In \emph{Proceedings of the 5th Workshop on Noisy User-generated
  Text, W-NUT@EMNLP}, pages 34--41.

\bibitem[{Liu et~al.(2015)Liu, Joty, and Meng}]{liu-etal-2015-fine}
Pengfei Liu, Shafiq Joty, and Helen Meng. 2015.
\newblock \href {https://doi.org/10.18653/v1/D15-1168} {Fine-grained opinion
  mining with recurrent neural networks and word embeddings}.
\newblock In \emph{Proceedings of the 2015 Conference on Empirical Methods in
  Natural Language Processing}, pages 1433--1443, Lisbon, Portugal. Association
  for Computational Linguistics.

\bibitem[{Manning et~al.(2014)Manning, Surdeanu, Bauer, Finkel, Bethard, and
  McClosky}]{manning-EtAl:2014:P14-5}
Christopher~D. Manning, Mihai Surdeanu, John Bauer, Jenny Finkel, Steven~J.
  Bethard, and David McClosky. 2014.
\newblock \href {http://www.aclweb.org/anthology/P/P14/P14-5010} {The
  {Stanford} {CoreNLP} natural language processing toolkit}.
\newblock In \emph{Association for Computational Linguistics (ACL) System
  Demonstrations}, pages 55--60.

\bibitem[{Mathew et~al.(2019)Mathew, Fakhraei, and
  Ambite}]{mathew2019biomedical}
Joel Mathew, Shobeir Fakhraei, and Jos{\'e}~Luis Ambite. 2019.
\newblock Biomedical named entity recognition via reference-set augmented
  bootstrapping.
\newblock \emph{arXiv preprint arXiv:1906.00282}.

\bibitem[{Mikheev et~al.(1999)Mikheev, Moens, and Grover}]{mikheev1999named}
Andrei Mikheev, Marc Moens, and Claire Grover. 1999.
\newblock Named entity recognition without gazetteers.
\newblock In \emph{Proceedings of the ninth conference on European chapter of
  the Association for Computational Linguistics}, pages 1--8. Association for
  Computational Linguistics.

\bibitem[{Miller(1995)}]{miller1995wordnet}
George~A Miller. 1995.
\newblock Wordnet: a lexical database for english.
\newblock \emph{Communications of the ACM}, 38(11):39--41.

\bibitem[{Plank et~al.(2016)Plank, S{\o}gaard, and
  Goldberg}]{plank-etal-2016-multilingual}
Barbara Plank, Anders S{\o}gaard, and Yoav Goldberg. 2016.
\newblock \href {https://doi.org/10.18653/v1/P16-2067} {Multilingual
  part-of-speech tagging with bidirectional long short-term memory models and
  auxiliary loss}.
\newblock In \emph{Proceedings of the 54th Annual Meeting of the Association
  for Computational Linguistics (Volume 2: Short Papers)}, pages 412--418,
  Berlin, Germany. Association for Computational Linguistics.

\bibitem[{Pontiki et~al.(2016)Pontiki, Galanis, Papageorgiou, Androutsopoulos,
  Manandhar, AL-Smadi, Al-Ayyoub, Zhao, Qin, De~Clercq, Hoste, Apidianaki,
  Tannier, Loukachevitch, Kotelnikov, Bel, Jim{\'e}nez-Zafra, and
  Eryi{\u{g}}it}]{pontiki-etal-2016-semeval}
Maria Pontiki, Dimitris Galanis, Haris Papageorgiou, Ion Androutsopoulos,
  Suresh Manandhar, Mohammad AL-Smadi, Mahmoud Al-Ayyoub, Yanyan Zhao, Bing
  Qin, Orph{\'e}e De~Clercq, V{\'e}ronique Hoste, Marianna Apidianaki, Xavier
  Tannier, Natalia Loukachevitch, Evgeniy Kotelnikov, Nuria Bel,
  Salud~Mar{\'\i}a Jim{\'e}nez-Zafra, and G{\"u}l{\c{s}}en Eryi{\u{g}}it. 2016.
\newblock \href {https://doi.org/10.18653/v1/S16-1002} {{S}em{E}val-2016 task
  5: Aspect based sentiment analysis}.
\newblock In \emph{Proceedings of the 10th International Workshop on Semantic
  Evaluation ({S}em{E}val-2016)}, pages 19--30, San Diego, California.
  Association for Computational Linguistics.

\bibitem[{Pontiki et~al.(2015)Pontiki, Galanis, Papageorgiou, Manandhar, and
  Androutsopoulos}]{pontiki-etal-2015-semeval}
Maria Pontiki, Dimitris Galanis, Haris Papageorgiou, Suresh Manandhar, and Ion
  Androutsopoulos. 2015.
\newblock \href {https://doi.org/10.18653/v1/S15-2082} {{S}em{E}val-2015 task
  12: Aspect based sentiment analysis}.
\newblock In \emph{Proceedings of the 9th International Workshop on Semantic
  Evaluation ({S}em{E}val 2015)}, pages 486--495, Denver, Colorado. Association
  for Computational Linguistics.

\bibitem[{Pontiki et~al.(2014)Pontiki, Galanis, Pavlopoulos, Papageorgiou,
  Androutsopoulos, and Manandhar}]{pontiki-etal-2014-semeval}
Maria Pontiki, Dimitris Galanis, John Pavlopoulos, Harris Papageorgiou, Ion
  Androutsopoulos, and Suresh Manandhar. 2014.
\newblock \href {https://doi.org/10.3115/v1/S14-2004} {{S}em{E}val-2014 task 4:
  Aspect based sentiment analysis}.
\newblock In \emph{Proceedings of the 8th International Workshop on Semantic
  Evaluation ({S}em{E}val 2014)}, pages 27--35, Dublin, Ireland. Association
  for Computational Linguistics.

\bibitem[{Raille et~al.(2020)Raille, Djambazovska, and Musat}]{raille2020fast}
Guillaume Raille, Sandra Djambazovska, and Claudiu Musat. 2020.
\newblock Fast cross-domain data augmentation through neural sentence editing.
\newblock \emph{arXiv preprint arXiv:2003.10254}.

\bibitem[{Sang and De~Meulder(2003)}]{sang2003introduction}
Erik~F Sang and Fien De~Meulder. 2003.
\newblock Introduction to the conll-2003 shared task: Language-independent
  named entity recognition.
\newblock \emph{arXiv preprint cs/0306050}.

\bibitem[{Schl{\"u}ter and Grill(2015)}]{schluter2015exploring}
Jan Schl{\"u}ter and Thomas Grill. 2015.
\newblock Exploring data augmentation for improved singing voice detection with
  neural networks.
\newblock In \emph{ISMIR}, pages 121--126.

\bibitem[{Sch{\"u}tze(1993)}]{schutze1993part}
Hinrich Sch{\"u}tze. 1993.
\newblock Part-of-speech induction from scratch.
\newblock In \emph{Proceedings of the 31st annual meeting on Association for
  Computational Linguistics}, pages 251--258. Association for Computational
  Linguistics.

\bibitem[{Sennrich et~al.(2016)Sennrich, Haddow, and
  Birch}]{sennrich-etal-2016-improving}
Rico Sennrich, Barry Haddow, and Alexandra Birch. 2016.
\newblock \href {https://doi.org/10.18653/v1/P16-1009} {Improving neural
  machine translation models with monolingual data}.
\newblock In \emph{Proceedings of the 54th Annual Meeting of the Association
  for Computational Linguistics (Volume 1: Long Papers)}, pages 86--96, Berlin,
  Germany. Association for Computational Linguistics.

\bibitem[{Serban et~al.(2016)Serban, Garcia-Duran, Gulcehre, Ahn, Chandar,
  Courville, and Bengio}]{serban2016generating}
Iulian~Vlad Serban, Alberto Garcia-Duran, Caglar Gulcehre, Sungjin Ahn, Sarath
  Chandar, Aaron Courville, and Yoshua Bengio. 2016.
\newblock Generating factoid questions with recurrent neural networks: The 30m
  factoid question-answer corpus.
\newblock In \emph{Proceedings of the 54th Annual Meeting of the Association
  for Computational Linguistics (Volume 1: Long Papers)}, pages 588--598.

\bibitem[{Shang et~al.(2018)Shang, Liu, Jiang, Ren, Voss, and
  Han}]{shang2018automated}
Jingbo Shang, Jialu Liu, Meng Jiang, Xiang Ren, Clare~R Voss, and Jiawei Han.
  2018.
\newblock Automated phrase mining from massive text corpora.
\newblock \emph{IEEE Transactions on Knowledge and Data Engineering},
  30(10):1825--1837.

\bibitem[{Simard et~al.(1998)Simard, LeCun, Denker, and Victorri}]{Simard1998}
Patrice~Y. Simard, Yann~A. LeCun, John~S. Denker, and Bernard Victorri. 1998.
\newblock \href {https://doi.org/10.1007/3-540-49430-8_13}
  {\emph{Transformation Invariance in Pattern Recognition --- Tangent Distance
  and Tangent Propagation}}, pages 239--274. Springer Berlin Heidelberg,
  Berlin, Heidelberg.

\bibitem[{Sundermeyer et~al.(2012)Sundermeyer, Schl{\"u}ter, and
  Ney}]{sundermeyer2012lstm}
Martin Sundermeyer, Ralf Schl{\"u}ter, and Hermann Ney. 2012.
\newblock Lstm neural networks for language modeling.
\newblock In \emph{Thirteenth annual conference of the international speech
  communication association}.

\bibitem[{Tjong Kim~Sang(2002)}]{tjong-kim-sang-2002-introduction}
Erik~F. Tjong Kim~Sang. 2002.
\newblock \href {https://www.aclweb.org/anthology/W02-2024} {Introduction to
  the {C}o{NLL}-2002 shared task: Language-independent named entity
  recognition}.
\newblock In \emph{{COLING}-02: The 6th Conference on Natural Language Learning
  2002 ({C}o{NLL}-2002)}.

\bibitem[{Tjong Kim~Sang and
  De~Meulder(2003)}]{tjong-kim-sang-de-meulder-2003-introduction}
Erik~F. Tjong Kim~Sang and Fien De~Meulder. 2003.
\newblock \href {https://www.aclweb.org/anthology/W03-0419} {Introduction to
  the {C}o{NLL}-2003 shared task: Language-independent named entity
  recognition}.
\newblock In \emph{Proceedings of the Seventh Conference on Natural Language
  Learning at {HLT}-{NAACL} 2003}, pages 142--147.

\bibitem[{Wang and Eisner(2016)}]{wang2016galactic}
Dingquan Wang and Jason Eisner. 2016.
\newblock The galactic dependencies treebanks: Getting more data by
  synthesizing new languages.
\newblock \emph{Transactions of the Association for Computational Linguistics},
  4:491--505.

\bibitem[{Wang et~al.(2019)Wang, Wang, and Lian}]{wang2019survey}
Xiang Wang, Kai Wang, and Shiguo Lian. 2019.
\newblock A survey on face data augmentation.
\newblock \emph{arXiv preprint arXiv:1904.11685}.

\bibitem[{Wei and Zou(2019)}]{wei2019eda}
Jason Wei and Kai Zou. 2019.
\newblock Eda: Easy data augmentation techniques for boosting performance on
  text classification tasks.
\newblock In \emph{Proceedings of the 2019 Conference on Empirical Methods in
  Natural Language Processing and the 9th International Joint Conference on
  Natural Language Processing (EMNLP-IJCNLP)}, pages 6383--6389.

\bibitem[{Weston et~al.(2015)Weston, Bordes, Chopra, Rush, van Merri{\"e}nboer,
  Joulin, and Mikolov}]{weston2015towards}
Jason Weston, Antoine Bordes, Sumit Chopra, Alexander~M Rush, Bart van
  Merri{\"e}nboer, Armand Joulin, and Tomas Mikolov. 2015.
\newblock Towards ai-complete question answering: A set of prerequisite toy
  tasks.
\newblock \emph{arXiv preprint arXiv:1502.05698}.

\bibitem[{Yarowsky et~al.(2001)Yarowsky, Ngai, and
  Wicentowski}]{yarowsky2001inducing}
David Yarowsky, Grace Ngai, and Richard Wicentowski. 2001.
\newblock Inducing multilingual text analysis tools via robust projection
  across aligned corpora.
\newblock Technical report, JOHNS HOPKINS UNIV BALTIMORE MD DEPT OF COMPUTER
  SCIENCE.

\bibitem[{Yasunaga et~al.(2018)Yasunaga, Kasai, and Radev}]{yasunaga2018robust}
Michihiro Yasunaga, Jungo Kasai, and Dragomir Radev. 2018.
\newblock Robust multilingual part-of-speech tagging via adversarial training.
\newblock In \emph{Proceedings of the 2018 Conference of the North American
  Chapter of the Association for Computational Linguistics: Human Language
  Technologies, Volume 1 (Long Papers)}, pages 976--986.

\bibitem[{Yu et~al.(2018)Yu, Dohan, Le, Luong, Zhao, and Chen}]{wei2018fast}
Adams~Wei Yu, David Dohan, Quoc Le, Thang Luong, Rui Zhao, and Kai Chen. 2018.
\newblock \href {https://openreview.net/forum?id=B14TlG-RW} {Fast and accurate
  reading comprehension by combining self-attention and convolution}.
\newblock In \emph{International Conference on Learning Representations}.

\end{thebibliography}
